\documentclass{article}

\PassOptionsToPackage{numbers}{natbib}

\usepackage[preprint]{neurips_2024}

\usepackage[utf8]{inputenc} 
\usepackage[T1]{fontenc}    
\usepackage{hyperref}       
\usepackage{url}            
\usepackage{booktabs}       
\usepackage{amsfonts}       
\usepackage{nicefrac}       
\usepackage{microtype}      
\usepackage{xcolor}         

\usepackage{amsmath}
\usepackage{comment}
\usepackage{colortbl}
\usepackage{lipsum}
\usepackage{graphicx}
\usepackage{subcaption}
\usepackage{algorithm}
\usepackage{algpseudocode}
\usepackage{multirow}
\usepackage{xspace}

\algnewcommand{\Inputs}[1]{%
  \State \textbf{Inputs:}
  \Statex \hspace*{\algorithmicindent}\parbox[t]{.8\linewidth}{\raggedright #1}
}
\algnewcommand{\Initialize}[1]{%
  \State \textbf{Initialize:}
  \Statex \hspace*{\algorithmicindent}\parbox[t]{.8\linewidth}{\raggedright #1}
}

\title{Learning Visual Prompts for Guiding the Attention of Vision Transformers}

\author{
  Razieh Rezaei\thanks{Equal Contribution} \\
  School of Computation, Information and Technology \\
  Technical University of Munich \\
  \And
  Masoud Jalili Sabet$^*$ \\
  AI Innovation, Data:Lab Munich \\
  Volkswagen AG \\
  \AND
  Jindong Gu\thanks{Corresponding Author} \\
  Department of Engineering Science \\
  University of Oxford \\
  \And
  Daniel Rueckert \\
  School of Computation, Information and Technology \\
  Technical University of Munich \\
  \And
  Philip Torr \\
  Department of Engineering Science \\
  University of Oxford \\
  \And
  Ashkan Khakzar \\
  Department of Engineering Science \\
  University of Oxford \\
 }

\def\clipb{CLIP-B/32\xspace}
\def\clipl{CLIP-L/14\xspace}
\def\siglip{SigLIP\xspace}
\def\deit{DeiT\xspace}
\def\dino{DINOv2\xspace}

\begin{document}

\maketitle

\begin{abstract}
Visual prompting infuses visual information into the input image to adapt models toward specific predictions and tasks. Recently, manually crafted markers such as red circles are shown to guide the model to attend to a target region on the image. However, these markers only work on models trained with data containing those markers. Moreover, finding these prompts requires guesswork or prior knowledge of the domain on which the model is trained. This work circumvents manual design constraints by proposing to learn the visual prompts for guiding the attention of vision transformers. The learned visual prompt, added to any input image would redirect the attention of the pre-trained vision transformer to its spatial location on the image. Specifically, the prompt is learned in a self-supervised manner without requiring annotations and without fine-tuning the vision transformer. Our experiments demonstrate the effectiveness of the proposed optimization-based visual prompting strategy across various pre-trained vision encoders.
\end{abstract}    
\section{Introduction}
\label{sec:intro}

The attention mechanism \cite{vaswani2017attention} in vision transformers leverages contextually relevant visual information from the entire input image to form embeddings for each token. This process dynamically adjusts the representation of each token, allowing the model to focus selectively on relevant areas, and integrates this context into the embeddings. Such context-aware embeddings lead to enhanced recognition of complex visual data \cite{dosovitskiy2021imageViT}.  In our study, we explore whether it is possible to direct the attention mechanism toward a specific input region by introducing and constructing a marker for pre-trained vision transformer models. We pursue this investigation to better \emph{understand what kind of visual information directs the attention mechanism} of various pre-trained vision transformers and to \emph{explore its potential for visual prompting}. 

Visual prompting ~\cite{gu2023systematic} refers to approaches that infuse visual information to the input images to adapt vision foundation models to new tasks. 
Recent works introduce manual prompting techniques, such as placing colored circle or square markers\cite{Shtedritski_2023_ICCV_RedCircle} on the image, or blurring regions surrounding the prompted location \cite{yang2024fineBlur}, 
and show they are effective in adapting CLIP vision encoder \cite{radford2021learningCLIP}
for fine-grained recognition tasks. These works hypothesize that these manual prompting techniques guide the attention of the model and therefore help the model with fine-grained recognition tasks.
However, using these manually engineering prompts rely on prior knowledge of biases (or emergent properties) formed in the model during training (It is reported that training data of CLIP contains these markers and blur effects \cite{Shtedritski_2023_ICCV_RedCircle,yang2024fineBlur}).

In this study, we propose \textit{learning} prompts (without fine-tuning the model) to guide the attention of various large models rather than manually designing them. We explore how we can find visual prompts that attract the attention of a pre-trained (and frozen) vision transformer (ViT~\cite{dosovitskiy2021imageViT}) through a self-supervised approach. \emph{Thus, we would not rely on prior knowledge regarding dataset biases, and we can find prompts for any vision transformer trained on any dataset}. The optimization procedure alleviates the need for manual engineering or intuition to find the prompt.
Drawing inspiration from universal adversarial patches \cite{luo2024worth1000lies},
and employing techniques to improve transferability such as utilizing a network prior ~\cite{ulyanov2018deepPrior} 
to generate adversarial perturbations~\cite{poursaeed2018generativeGAP}, we propose a self-supervised prompt optimization pipeline.
In addition to CLIP, we explore other ViT networks, such as \deit~\cite{touvron2021trainingDeiT} and \dino~\cite{oquab2023dinov2,caron2021emergingDINO}, 
each of which follows different training principles than CLIP (e.g. CLIP uses language supervision, while \dino uses self-supervision). 
Identifying visual prompts for \dino is particularly important because it extracts significantly richer visual features from images and is recently emerging in the new generation of vision-language models \cite{tong2024eyesWide,kar2024brave} due to CLIP's visual limitations ~\cite{tong2024eyesWide}.
We observe that manual markers, like a red circle, do not effectively guide \dino's attention, highlighting the need for an optimization-based approach for finding visual prompts.

\begin{figure*}
\begin{center}
\includegraphics[width=\textwidth]{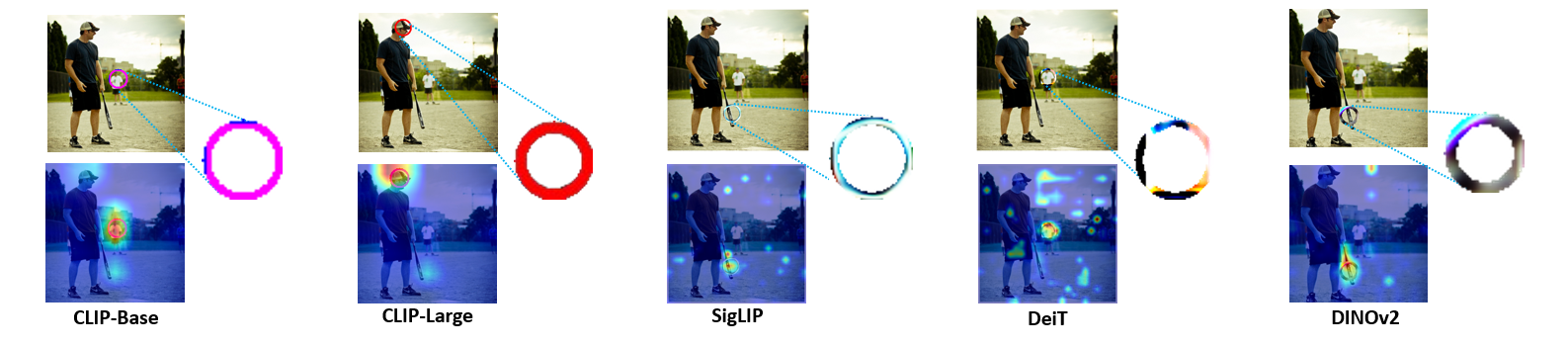}
\caption{\textbf{Learned Prompt for CLIPs \cite{radford2021learningCLIP}, \siglip~\cite{zhai2023sigmoid}, \deit~\cite{touvron2021trainingDeiT}, \dino~ \cite{oquab2023dinov2}.} In our framework, we learn a prompt to draw the attention of the model to a specific point where the prompt is applied. The prompt is optimized for each vision encoder model specifically to generalize across different images. The depicted image is taken from COCO~\cite{lin2014microsoftCoco}. 
}
\label{fig:overview}
\end{center}
\end{figure*}

\section{Related Work}
\label{sec:RW}

\noindent\textbf{Visual Prompting}
Prompting has been extensively studied in the NLP community~\cite{liu2023pre}. Recently, researchers have begun exploring the benefits of prompting in image recognition as well. This involves adding a learnable modification to images to guide models towards specific predictions~\cite{zhou2022learning,jia2022visualpromptTuning,bahng2022exploringVisualPromptIsola,shen2024multitaskVisualTuningDarell,gu2023systematic}. 
These prompt-tuning methods ~\cite{jia2022visualpromptTuning,bahng2022exploringVisualPromptIsola,shen2024multitaskVisualTuningDarell,zhou2022learning} optimize a visual prompt that is appended and \emph{fixed} to an input image.
For instance, this could involve appending optimizable pixel regions around the image.
By doing so, these methods enable the pre-trained model to adapt to new tasks without the need for retraining.
This modification is typically applied universally across different images.
Follow up approaches have been proposed to improve these visual prompts~\cite{zhou2022conditional,zang2022unified,khattak2023maple}.
More recently, researchers have shown that manually crafted prompts, like a red circle, can effectively guide the attention of models like CLIP when trained on datasets containing similar markers
~\cite{Shtedritski_2023_ICCV_RedCircle}.
Moreover, other biases, such as blurring, can be used for prompting ~\cite{yang2024fineBlur}. However, such attention-guiding prompts only work effectively on certain models. In this study, we propose learning prompts (without fine-tuning the model) to guide the attention of various models rather than manually designing them.

\vspace{0.1cm}
\noindent\textbf{Adversarial Patch}
The seminal work~\cite{papernot2016limitations} demonstrates that adversarial examples can be generated by altering just a few pixels in the input. This white-box attack~\cite{karmon2018lavan,liu2019perceptual,wang2021universal,luo2021generating} method can deceive models by targeting a very small area.
Additionally, certain studies~\cite{brown2017adversarial,liu2020bias} have successfully developed universal, transferable, and targeted adversarial patches.
These patches are typically placed on the primary object within images.
A related study~\cite{gu2022vision} found that model attention can be drawn towards adversarial patches designed to mislead classification results.
Our visual prompt can similarly be viewed as an adversarial patch, aiming to divert model attention away from its original focus, but for adapting model behavior towards useful tasks.

\vspace{0.1cm}
\noindent\textbf{Transferability and Universality of Adversarial Perturbation} The transferability of adversarial perturbation describes the phenomenon that perturbations crafted for one model can deceive another, even with a different architecture~\cite{gu2023survey}.
Different from the transferability, the universality of adversarial perturbation is a property that the perturbation is still effective when it is added to different input instances~\cite{chaubey2020universal}. 
To improve transferability, we use generative adversarial perturbations ~\cite{poursaeed2018generativeGAP}.
A simple way to increase the universality is to include various input images when optimizing an adversarial perturbation~\cite{moosavi2017universal}, which we also leverage in this work.
The transferability and universality of adversarial perturbation have also been studied~\cite{brown2017adversarial,liu2020bias,xiao2021improving}.
These intriguing properties increase the threats in real-world applications.
However, how to leverage the two properties for good has not been explored in the community.

\begin{figure*}
\begin{center}
\includegraphics[width=\textwidth]{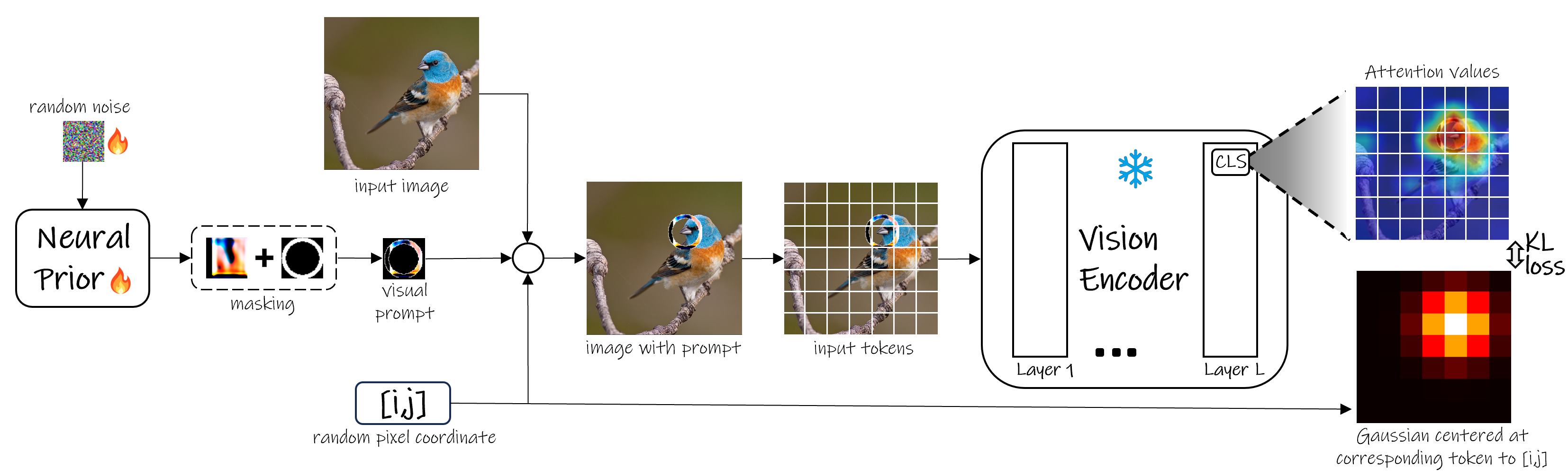}
\caption{\textbf{Overview of Self-supervised Prompt Optimization Framework:} 
For a given image and a random position for the patch, the patch prior (a random noise) is first passed through a neural prior (an auto-encoder neural network).
A desired mask is then applied to the patch to only partially cover the image and avoid losing data.
The prompt is positioned on the target location and, after passing through the frozen Vision Encoder, the attention weights are extracted.
The desired attention target values are calculated with a Gaussian distribution centered at the token corresponding to the target location.
During training, the framework learns a prompt that minimizes the Kullback-Leibler (KL) divergence loss between the attention values of the CLS token and the target distribution.
}
\label{fig:method_overview}
\end{center}
\end{figure*}

\section{Method}
\label{sec:formatting}

Our goal is to learn a patch such that if it is applied to any part of an input image, the corresponding tokens in the vision encoder would attract more attention, and therefore, the final representation of the image will be manipulated.
We propose a self-supervised method to learn such a patch, while the vision-encoder model is utilized in its frozen state and we only require a collection of images without the need for any labels.
The learning is about the patch's RGB color space (three channels) while its shape is predefined, or alternatively, the shape mask can be learned as an additional fourth channel.
In the subsequent sections, we provide a detailed explanation of how this prompt is learned through the back-propagation of the self-supervised loss that accounts for the relation between the patch's location and the attention it receives.

\textbf{Input with Prompt.}
We aim to discover a visual prompt (patch) that attracts the attention of a transformer-based vision encoder toward a specific location.
To achieve this, we work on the RGB image input denoted as $I \in \mathbb{R}^{n \times n \times 3}$ where $n$ is the number of pixels in one dimension of the image and $3$ indicates the three-dimensional color space.
The prompt, denoted as $\mathcal{P}$, is also in a similar RGB color space and of size $m$$\times$$m$ (the impact of prompt size on performance is investigated in section 4.1).
The designated coordinates $[i, j]$ within the image $0 \leq i,j < n$ is then the exact pixel location where the visual prompt will be centrally inserted on.
By selecting coordinates only with criteria of $0 < i\pm \frac{m}{2}, j\pm \frac{m}{2} < n$, we ensure that the patch falls within the valid range, meaning that no part of the patch extends beyond the boundaries of the image after it is inserted.

The key point here is that by \textit{inserting} we do not simply mean adding the values of the prompt pixel with those of the image pixel.
Rather, we aim to identify a general, universal patch that remains effective when applied to different images and locations, regardless of the underlying pixel values.
To achieve this, we \textit{replace} the values of the corresponding image pixels with those of the patch pixels 
$I[i-\frac{m}{2}:i+\frac{m}{2}, j-\frac{m}{2}:j+\frac{m}{2}]=\mathcal{P}[:,:]$,
a process we refer to as \textit{insertion} of patch into the image.
To ensure the transferability of the visual prompt, the placement of the patch is randomized across various locations within the image.
During the training phase, for an input image, we select \( k \) random locations denoted by coordinates \([i, j]\) within the validity range criteria mentioned.
The patch is then inserted on each coordinate one at a time.
This process is repeated for all images $\mathcal{I}$ in the dataset, resulting in a set of modified input images $I_{\mathcal{P}}$ for all samples.
These modified images, with patches inserted at random locations, are then fed into the vision encoder.

\textbf{Transformer}\textrm{-}\textbf{based Vision Encoder.}
Within a transformer block with \( L \) layers and \( H \) attention heads, we denote the attention values as \( A^i_l \in \mathbb{R}^{t \times t} \), where \( A \) is the attention values of the \( i^{th} \) head of the layer \( l \) for an input image with \( t \) tokens.
The output of the transformer encoder is typically a contextualized representation of the image tokens, where each token has been influenced by its surrounding tokens through the attention mechanism.
This representation, often obtained from a special token (e.g., the CLS token), can then be utilized for downstream vision tasks such as object recognition or image classification.
Our method leverages the attention values of the CLS token in the last layer, averaged over the heads of that layer: \( \bar{A}_{L}[CLS,*]=\sum_{i=1}^{H}A^i_{L}[CLS,*] \).

\textbf{Target Gaussian Map.} 
When the prompt is applied to the input image on the pixel position $[i,j]$, it overlays on one or more tiles with the center $[x,y]$, refer to
Fig.~\ref{fig:method_overview}.
Assuming that the model divides an $n\times n$ pixel image to $t\times t$ image tokens with each tile having the pixel size $n_t$ ($n / t = n_t$),
the corresponding patch center $[x,y]$ can be derived from its pixel position on the image $[i,j]$ as: $x,y = (i/n_t, j/n_t)$.

To construct the target Gaussian map for the patch over the $t\times t$ token space, a 2D Gaussian distribution is employed.
The Gaussian map $\mathcal{G}(x, y)$ at location $[x,y]$ is represented by the probability density function of a Gaussian distribution $N(\mu ,\sigma ^{2})$, where $\mu=(x,y)$ denotes the mean (center) of the distribution,
and $\sigma$ is calculated from the Full Width at Half Maximum (FWHM), which in our experiments is set to the patch size in token space $m / n_t$: $\sigma = (m / n_t) / (2 \sqrt{2 \ln(2)})$.


\textbf{Neural Prior.}
To learn the optimized prompt, we aim to avoid the expensive need to fine-tune the large-scale vision encoder by employing it solely in a frozen state.
In this way, the only parameters being updated through the training process are the pixel values of the prompt. However, a potential barrier we may encounter in efficiently achieving the optimal patch is the limited learnable variables within the pixel space \cite{ulyanov2018deepPrior,Khakzar_2022_CVPR}.
Thus we parameterize the patch input space with a neural network as in \cite{ulyanov2018deepPrior}. Such parameterization is also shown to improve the transferability of optimized perturbations \cite{poursaeed2018generativeGAP}.
we use a neural prior $f$ starting from randomly initialized weights, that receives a random prior input $\eta \in m \times m$ , and outputs an initial prompt $\mathcal{P}_{prior} = f(\eta)$ which is then masked by a predefined shape mask 
$\mathcal{P} = \mathcal{P}_{prior} * \mathcal{P}_{mask}$ 
and finally inserted centrally on the $[i,j]$ pixel of the input image $I$ to form the input to the vision encoder: $ViT(I_\mathcal{P})$.
Our neural prior $f$ employs a CNN-based architecture, in particular a U-Net with three layers of downsampling and three upsampling with 
Sigmoid activation to ensure the values stay between $[0, 1]$.
Leveraging shared spatial patterns among pixels in the patch enables effective optimization during training which facilitates efficient visual prompt learning \cite{Krull_2019_CVPR}.

\begin{algorithm}[ht]
    \centering
    \caption{Learning Prompt with Predefined Shape}\label{algorithm}
    \begin{algorithmic}[1]
        \Require $\mathcal{I}$, $\eta$, $n_{t}$ \Comment{Image Collection, Patch Prior Noise, $n_{t}^{2}$ = pixel size of a ViT tile}
        \Initialize{$n = \textproc{pixel\_size}(I)$ \Comment{Image size: $n \times n$}}
        \For{$I \in \mathcal{I}$} 
            \State $\mathcal{P} \gets f_{\theta}(\eta)$ \Comment{Generate patch from patch prior}
            \State $[i,j] \gets \textproc{Random}(n)$ 
            \State $x,y \gets i/n_{t}, j/n_{t}$ \Comment{Find corresponding token coordinates}

            \State $I_p \gets \textproc{insert}(I, P, [i,j])$ 
            \State $\bar{A}_{CLS} \gets \textproc{Attention}(\textproc{ViT}(I_p))$ \Comment{Get the averaged attention values for CLS over all heads}

            \State $l \gets \mathcal{L}_{KL}(\bar{A}_{CLS}, \textproc{GaussMap}(x,y))$ 
            \State $\theta \gets \textproc{AdamW}(\theta, l)$ \Comment{Update the model parameters using AdamW optimizer}
        \EndFor

    \end{algorithmic}
\end{algorithm}


\textbf{Objective Function.}
Assuming that an image with a patch applied on it $I_{\mathcal{P}}=Insert(I, \mathcal{P}, [i,j])$ is the input to the $ViT$, we define mean of last-layer attention values over attention heads of the vision encoder model as $\bar{A}_{l}=Attention(ViT(I_{\mathcal{P}}))$.
Our objective is to train the deep neural prior to output a prompt $\mathcal{P}$ such that it enhances the attention $\bar{A}_{l}$ at corresponding token at $x,y=(i/n_t, j/n_t)$ with $n_{t} \times n_{t}$ indicating the number of pixels in a token.
Having this objective in mind, we calculate the final self-supervised loss as follows:
\begin{equation}
\mathcal{L}(I, [i,j]) = \mathcal{L}_{\text{KL}}(\bar{A}_{L}[CLS,*], \mathcal{G}(x,y))
\end{equation}
where $\mathcal{L}_{\text{KL}}$ is the KL-Divergence loss, Throughout the training process, the encoder remains frozen, and only the weights of the neural prior undergo updates as a result of the back-propagating of the defined self-supervised loss.

\section{Experiments}
In the following, we first present the learned prompts for \clipb and \clipl, discussing several design considerations and their impacts.
We then evaluate the effectiveness of the prompts in the context of Naming Keypoints task. We take it a step further by learning prompts for \siglip \cite{zhai2023sigmoid}, \deit~\cite{touvron2021trainingDeiT}, and \dino~\cite{oquab2023dinov2},
each of which either features a slightly different encoder architecture, or has a different training objective function, or varies in pretraining datasets.
Finally, we compare how the prompts impact their attention through layers.

\textbf{Implementation Details and Settings.}
All the vision encoders used are pretrained models available from publicly accessible libraries such as Hugging Face Transformers \cite{wolf2019huggingface} and PyTorch Hub \cite{paszke2017automatic}.
Unless specified, we use a learning rate of 1e-3 and batch size of 32 and train our framework for 10 epochs.
All experiments were conducted on a machine equipped with a single NVIDIA A100 GPU with 80GB of memory.

\textbf{Datasets and Evaluation Metric.}
\textit{Training:} To train our neural prior, we use images from ImageNet \cite{deng2009imagenet} which consists of millions of labeled images spanning one thousand classes.
However, as working with the full ImageNet is resource-demanding, we exploit a subset of it comprising of 10 
categories according to \cite{jph00}.
The limited coverage of label classes is not an issue for our work, as our method is not supervised and thus does not rely on specific label types.
Furthermore, the over 13k samples in \cite{jph00} offer a rich and diverse set of images, enhancing the generalizability of our self-supervised training.
\textit{Evaluation:} For naming a keypoint task, we utilize CUB-200-2011 test set \cite{WahCUB_200_2011} to evaluate our learned prompt.
This dataset contains 5794 bird images of different types, annotated with 15 body-part names and corresponding pixel coordinates as keypoints on the bird image.
We evaluate our method by positioning our learned prompt on these keypoints and then calculating the performance based on the percentage of correctly identified body parts.
To further evaluate our learned prompt, we test its performance also on the RefCOCO \cite{kazemzadeh2014referitgame}, RefCOCO+\cite{kazemzadeh2014referitgame}, and RefCOCOg\cite{mao2016generation}
datasets which consist of images, annotated with bounding boxes around objects in it, each of which is paired with expressions.

\textbf{Models and Baselines.}
Specifically, \clipb and \clipl are employed for their robust zero-shot learning capabilities, leveraging large-scale vision-language pretraining. 
\siglip integrates image and language understanding in a synergistic manner.
\deit and \dino are utilized for their state-of-the-art performance in vision transformer architectures, with \deit focusing on efficient training and \dino providing self-supervised learning benefits.
The baseline for comparison involves using the cropped region over the object bounding box in the RefCOCO dataset family.
This approach isolates the relevant object from the surrounding context, allowing for a focused evaluation of keypoint identification accuracy.
Additionally, random location baselines are employed to assess the robustness of our method against random visual prompts.

\subsection{Guiding Attention to Name Keypoints}

We initiate our experiments with a basic prompt configuration, termed a vanilla patch, which is a simple filled square.
This patch undergoes no shape mask filtering during training, and the learned prompt is optimized RGB values.
We train patches of varying sizes for the \clipb and \clipl models.
Our experiments utilize prompt sizes corresponding to the model token dimensions.
As CLIP models split input images into tokens of different pixel sizes (base model: $32 \times 32$, large model: $14 \times 14$), our learned prompt of one-token size inherently possesses varying pixel dimensions.

Table \ref{tab:vanilla} presents the visualizations of the prompts learned for both models at different sizes.
We select sizes proportional to the tokens: 1, 2, 3, and 4 times larger for \clipl.
For \clipb, due to the substantial token pixel size, we opt for prompts that are 1, 1.5, 2, and 2.5 times larger to avoid excessive pixel coverage by the patch.
We report the accuracy of keypoint naming on the CUB dataset.
The results indicate that for \clipb, the prompt size equal to 1 token yields the best performance, whereas for \clipl, the 3-times larger prompt achieves the highest accuracy.
Looking into this deeper, this suggests that optimal accuracy is achieved when the prompt covers approximately the same area size on the image ($32 \times 32$ and $42 \times 42$) for both models.
This may be because a larger patch, while better at manipulating the image, also covers more of it, leading to information loss and lower performance.


\begin{table}[!t]
\begin{center}
\caption{\textbf{Vanilla Patches and Scale Constraints.} The tile sizes of the \clipb and \clipl models are $32 \times 32$ and $14 \times 14$, respectively. We defined the dimensions for the vanilla patch (a simple filled square prompt) accordingly. For \clipb, the patch sizes are scaled proportionally to the model's input token at $[1, 1.5, 2, 2.5]$ times larger. For \clipl, the patch sizes are $[1, 2, 3, 4]$ times larger than the token. These scales are chosen to ensure the patch is sufficiently large for effective learning without excessively covering image information. These prompts are evaluated on the keypoint naming task using the CUB dataset.}\label{tab:vanilla}
\begin{tabular}{ |l|c|c|c|c|c|c|c|c| }
\cline{2-9}
\multicolumn{1}{c|}{} & \multicolumn{4}{c|}{(a) \clipb} & \multicolumn{4}{c|}{(b) \clipl}  \\
\hline
Prompt Size   & 32x32 & 48x48 & 64x64 & 80x80 & 14x14 & 28x28 & 42x42 & 56x56 \\ \hline
Visualization & 
\includegraphics[width=5mm, height=5mm]{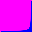} & 
\includegraphics[width=5mm, height=5mm]{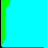} & 
\includegraphics[width=5mm, height=5mm]{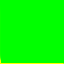} & 
\includegraphics[width=5mm, height=5mm]{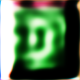} &
\includegraphics[width=5mm, height=5mm]{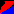} & 
\includegraphics[width=5mm, height=5mm]{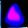} & 
\includegraphics[width=5mm, height=5mm]{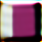} & 
\includegraphics[width=5mm, height=5mm]{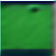} \\ \hline
CUB accuracy     & 10.97 & 9.85 & 9.71 & 7.35 & 12.86 & 16.04 & 17.41 & 13.68 \\ \hline
\end{tabular}

\end{center}
\end{table}

\begin{figure}[!t]
\begin{minipage}[t]{0.99\textwidth}
\centering
         \centering
     \begin{subfigure}[b]{0.21\textwidth}
         \centering
         \includegraphics[width=\textwidth]{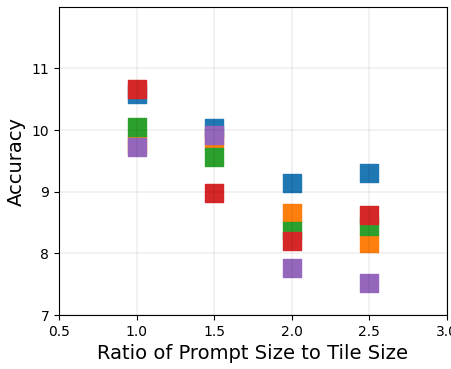}
         \caption{\clipb. Square}
     \end{subfigure}
     \hfill
     \begin{subfigure}[b]{0.21\textwidth}
         \centering
         \includegraphics[width=\textwidth]{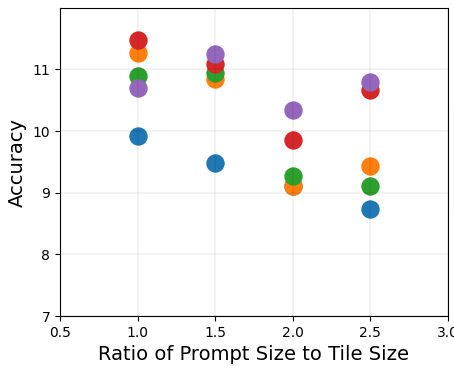}
         \caption{\clipb. Circle}
     \end{subfigure}
     \hfill
     \begin{subfigure}[b]{0.21\textwidth}
         \centering
         \includegraphics[width=\textwidth]{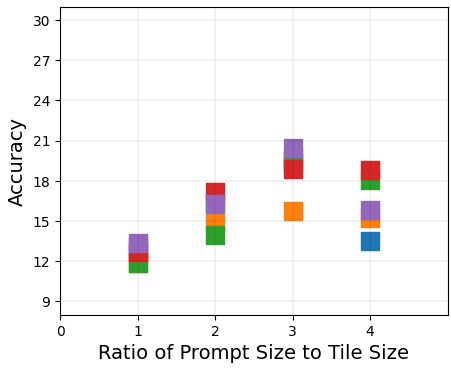}
         \caption{\clipl. Square}
     \end{subfigure}
     \hfill
     \begin{subfigure}[b]{0.21\textwidth}
         \centering
         \includegraphics[width=\textwidth]{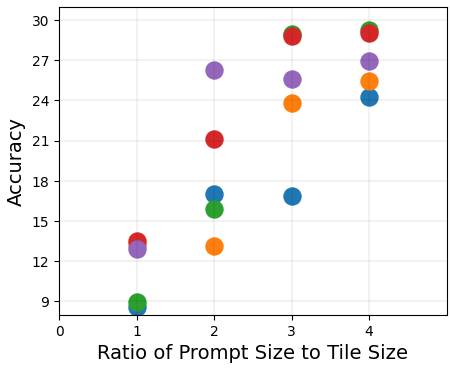}
         \caption{\clipl. Circle}
     \end{subfigure}
     \hfill
     \begin{subfigure}[b]{0.13\textwidth}
         \centering
         \includegraphics[width=\textwidth]{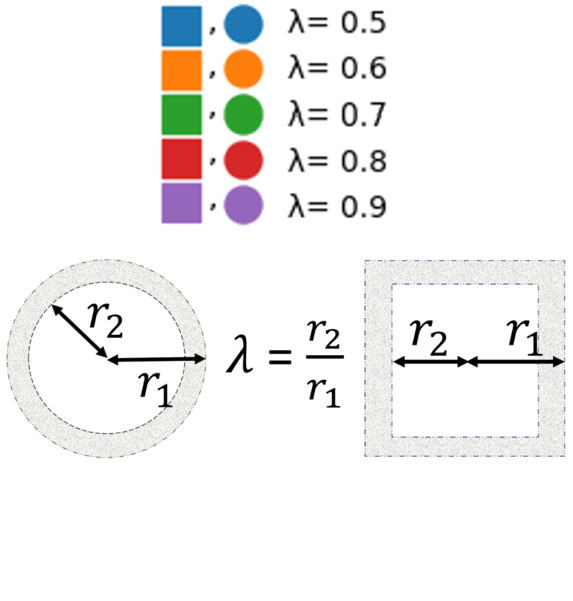}
         \caption{Lambda}
     \end{subfigure}
\end{minipage}
\hfill
\caption{\textbf{Effect of Patch Parameters:} 
Performance of CLIP models on CUB dataset for varying ratios of hollow circle and square masks over different patch sizes. 
The tile size of models ViT-Base and ViT-Large are $32$x$32$ and $14$x$14$, respectively. (e) shows the ratio colors and the illustration of $\lambda$.
}\label{fig:square_circle}
\end{figure}

\textbf{Beyond vanilla patch towards using shape priors: } Initially, we employed a filled square, but previous studies imply that the geometric shape of the patch influences its effectiveness
\cite{Shtedritski_2023_ICCV_RedCircle, bahng2022exploringVisualPromptIsola}.
To investigate the influence of prompt shape on performance in naming keypoints task for CLIP models, we employ hollow square and circle as predefined shape masks.
We define a thickness factor as $\lambda =$ $\frac{inner~Diameter}{outer~Diameter}$ as shown in Figure \ref{fig:square_circle} and investigate the performance for different thicknesses together with different patch sizes, again proportional to the model's token size.


What we observe from the performances in Figure \ref{fig:square_circle} is that the circle-shaped prompt yields higher accuracy compared to the square-shaped prompt. For \clipb, the circle prompt achieves the highest accuracy of 11.51, and the best square prompt yields slightly lower accuracy. For \clipl, the difference is significant: the circle prompt reaches an accuracy of approximately 30.5, whereas the square prompt only achieves about 20.5. The comparison of patch sizes reveals findings consistent with the vanilla patch results. For \clipb, the optimal patch size remains the same as the default size for one token. In contrast, for \clipl, the highest accuracy is obtained with prompt three times larger than one token of the large clip model, the size covering a similar number of pixels. The visualizations of the learned prompts with the highest performance are depicted in Table \ref{tab:visual_hollows}. We observe that the red color for \clipl and the pink color for \clipb are predominantly present in the learned prompts. Interestingly, for \clipl, when the prompt shape is a circle, it learned a solid red color which confirms the findings in \cite{Shtedritski_2023_ICCV_RedCircle}.

\begin{table}[!h]
\begin{center}
\caption{\textbf{Visualization of optimal prompt for various CLIP encoders.} Here we visualize the best-performing square frame and circle-shape prompts from Figure \ref{fig:square_circle}. C stands for Circle and S for Square. Colors of red and pink are predominant for \clipb and \clipl, respectively. It is noteworthy that the optimal prompt for \clipl is a red circle. Even when we use a square shape prior, a red circle emerges inside the square. Previously, \cite{Shtedritski_2023_ICCV_RedCircle} tried various markers using intuition and identified a red circle as an effective prompt. Interestingly, the marker seems to be the optimal marker for guiding the attention of \clipl. Though for \clipl the red circle is not optimum. Thus it may be an emergent property arising from scale.}\label{tab:visual_hollows}
\begin{tabular}{ |l|c|c|c|c| }
\cline{1-5}
Encoder/Shape  & \clipb, C & \clipb, S  & \clipl, C & \clipl, S \\  \hline
\rule{0pt}{5.5mm}
visualization
& \includegraphics[width=5mm, height=5mm]{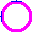}
& \includegraphics[width=5mm, height=5mm]{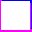}
& \includegraphics[width=5mm, height=5mm]{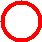}
& \includegraphics[width=5mm, height=5mm]{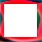} \\
\hline
\end{tabular}
\end{center}
\end{table}

Now that we have developed our prompt and explored the design configurations, we are interested in evaluating its performance compared to baseline methods on different datasets.
To this end, we train our method on the ImageNet dataset and test it on the CUB and RefCOCO datasets. These datasets are new to the prompt and have not been seen during the training phase.
Additionally, they offer the advantage of annotated image locations with names of specific body parts or objects for CUB and RefCOCO, respectively.

We define the baselines as \textit{Random}: Randomly selected areas. 
\textit{Crop}: Removing the area around the bounding box for RefCOCO and defining a square similar to the prompt size for CUB, then cutting out that area.
\textit{Blur}: Similar to cropping, but with the additional step of blurring the outer area using the averaging method with a kernel size of 5. 

As indicated by the results presented in Table \ref{tab:baslineExample}, the Crop method exhibits superior performance in RefCOCO tasks, but its effectiveness diminishes when applied to the CUB dataset.
In contrast, the Blur method shows notable improvements in RefCOCO tasks.
Our observations suggest that RefCOCO object descriptions often focus solely on the target object rather than relying heavily on scene context.
This tendency explains the success of both the Blur and Crop methods in this dataset.

However, in the CUB dataset, accurately recognizing animals can be challenging for the vision encoders, making it difficult to detect specific body parts.
Despite this challenge, our learned prompts generally outperform other baselines in the CUB dataset.
In RefCOCO tasks, our method consistently achieves second place after the effective Crop method.
These findings underscore the effectiveness of our approach, especially when the context of the image is important.

\begin{table}[t]
\centering
\caption{Performance comparison of different methods across various datasets. Bold numbers indicate the best performance, while underlined numbers denote the second-best performance.
The Crop method demonstrates superior performance in RefCOCO tasks but exhibits reduced effectiveness in the CUB dataset.
Conversely, the Blur method performs significantly better in RefCOCO tasks.
Our observations reveal that object descriptions in RefCOCO do not always rely heavily on scene context, explaining the success of both Blur and Crop methods.
However, in the CUB dataset, where context is crucial for proper animal detection, losing context leads to decreased model performance.}
\label{tab:baslineExample}
\begin{tabular}{|l|l|c|c c c|c c c|c c|}
\hline
\multirow{2}*{Encoder} & \multirow{2}*{Prompt}  &CUB& \multicolumn{3}{c|}{RefCOCO} & \multicolumn{3}{c|}{RefCOCO+} & \multicolumn{2}{c|}{RefCOCOg} \\
\cline{3-11}
 &  & K2N & TestA & TestB & Val & TestA & TestB & Val & Test & val \\
\hline
\hline
  & Random & 8.2 & 36.9 & 39.5 & 37.8 & 36.9 & 39.7 & 37.9 & 39.5 & 39.1 \\ \hline
\multirow{3}*{\scriptsize \textbf{\clipb}} & Crop & \textbf{15.6} & 56.6 & 47.7 & 52.6 & 60.4 & 51.7 & 55.9 & 60.3 & 59.5 \\
 & Blur & 7.4  & \underline{39.6} & \underline{41.1} & \underline{40.0} & \underline{39.5} & \underline{41.2} & \underline{40.0} & \underline{41.3} & \underline{41.1} \\
 & \includegraphics[width=4mm, height=4mm]{Figs/best_patches/patch_best_B_ci08.png}{\tiny $32\times 32$}
 &  \underline{11.39} & 37.7   & 40.9 & 39.3 & 38.5 & 40.2 & 38.5 & 39.1 & 39.0 \\
\hline
\multirow{3}*{\scriptsize \textbf{\clipl}} & Crop & 18.5 & 58.8 & 47.8 & 53.5 & 63.5 & 51.4 & 57.5 & 61.2 & 60.6 \\
 & Blur & 11.0  & 42.5 & 41.6 & 41.2 & 43.3 & 41.2 & 41.9 & 42.9 & 43.2 \\
 & \includegraphics[width=4mm, height=4mm]{Figs/best_patches/patch_best_L_ci08.png}{\tiny $42\times 42$}
 & \textbf{29.7} & \underline{49.5} & \underline{43.9} & \underline{46.7} & \underline{51.3} & \underline{46.2} & \underline{48.1} & \underline{50.4} & \underline{49.0} \\
\hline
\multirow{3}*{\scriptsize \textbf{\siglip}} & Crop & 19.6 & 62.6 & 51.3 & 57.9 & 66.8 & 56.7 & 62.3 & 69.2 & 68.5 \\
 & Blur & 13.1  & 42.9 & 41.3 & 41.8 & 43.6 & 43.0 & 42.8 & 44.2 & 44.5 \\
 & \includegraphics[width=4mm, height=4mm]{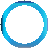}{\tiny $48\times 48$}
 & \textbf{30.8} & \underline{43.7} & \underline{43.5} & \underline{42.9} & \underline{45.1} & \underline{43.8} & \underline{44.0} & \underline{45.1} & \underline{45.0} \\
\hline
\end{tabular}
\end{table}

\subsection{Investigating Attention Dynamics in Various Vision Encoders}

Various vision encoder models are trained on diverse datasets, leading to behaviors that may differ from CLIP models.
More importantly, these vision foundations use different training paradigms.
CLIP vision encoder is trained with language supervision, while \dino is trained by self-supervision using visual information only.
Therefore, during training is no supervision signal for the model to learn that colored markers (such as red circles) signify the importance of the region the marker is placed.
Additionally, their architectural details might vary slightly, e.g., \siglip has no CLS token and instead averages over tokens by an attention pooling mechanism.
As a result, the optimal prompt to redirect their attention can also be different.
To investigate this, we use a similar self-supervised training approach to learn prompts of \siglip, \deit, and \dino as well.

\begin{figure}[t]
\begin{center}
\vskip -5pt
\includegraphics[height=2.6in, width=0.99\textwidth, keepaspectratio]{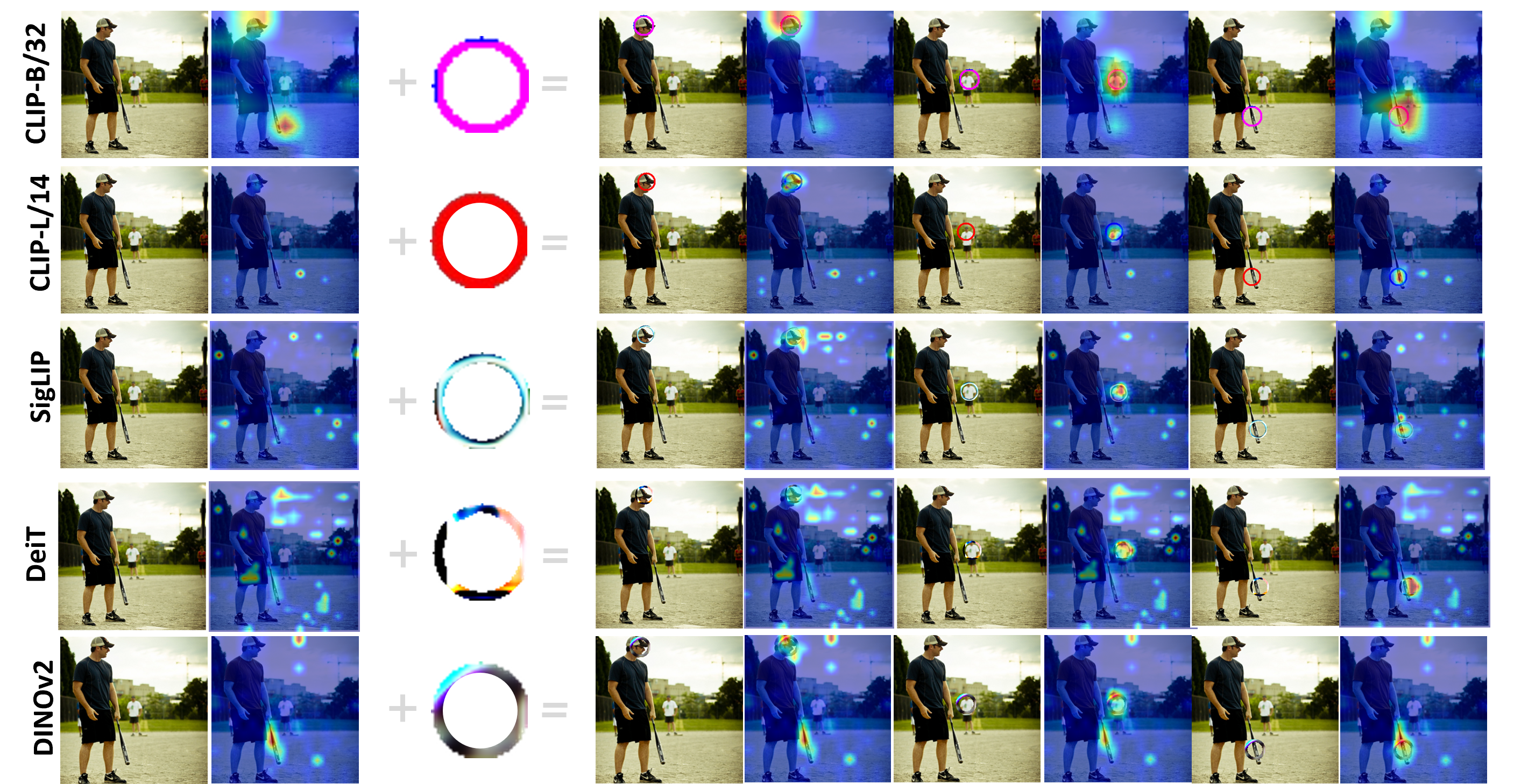}
\caption{\textbf{Learned Prompt for Different pretrained Vision Transformers.} The optimal visual prompts are not the same and each model has its own unique pattern. The prompt is optimized to generalize across images. Comparing attention heatmaps of the original (on left) and prompted images (on right) reveals how effectively the prompt directs attention to specific locations. The prompt is optimized for each ViT (\clipb, \clipl, \siglip, \deit, and \dino) separately, and is optimized over 20k random samples from ImageNet~\cite{deng2009imagenet}. The depicted image is taken from MS COCO~\cite{lin2014microsoftCoco}.}
\label{fig:prompt_vis_for_all}
\end{center}
\end{figure}

Figure \ref{fig:prompt_vis_for_all} shows the visualization of the learned prompt for each vision encoder. The different appearance of the visual prompts confirms that there is no single prompt that is universally optimal for all encoders.
We apply the learned prompts on to three different locations on an image and compare the heatmap visualizations of the prompted image with the original, unprompted image.
For almost all encoders, the attention heatmap shifts towards the prompt location, demonstrating the effectiveness of our prompt in manipulating the attention of the corresponding visual encoder models.

\begin{figure}[!t]
     \begin{subfigure}[b]{0.18\textwidth}
         \centering
         \includegraphics[width=\textwidth]{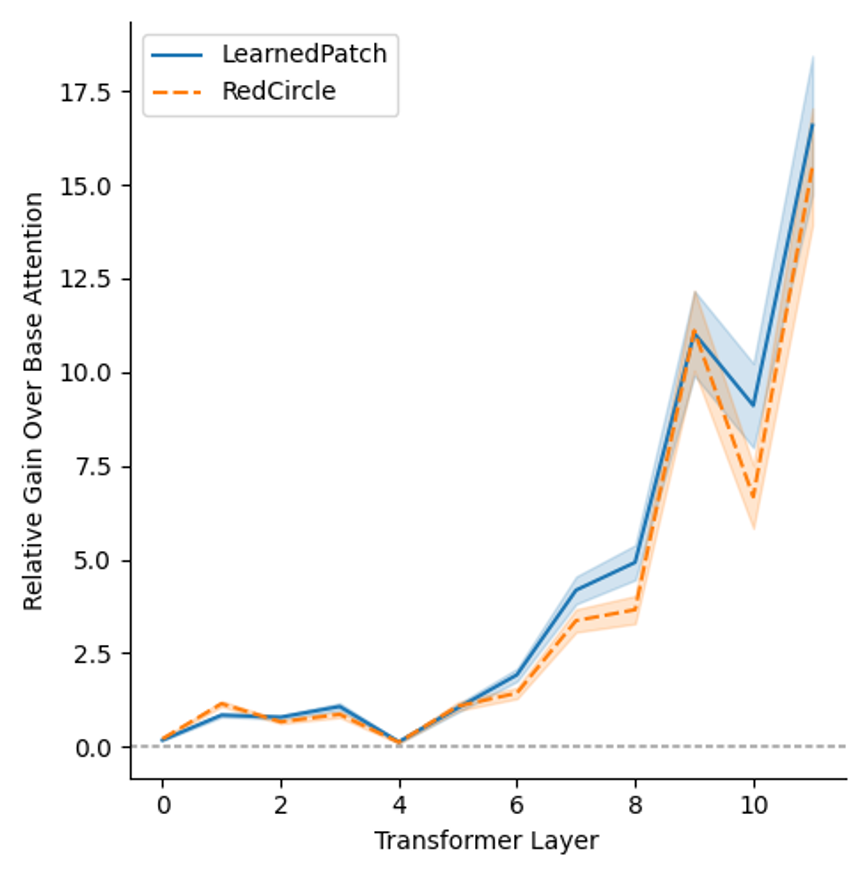}
         \caption{\clipb}
     \end{subfigure}
     \hfill
     \begin{subfigure}[b]{0.18\textwidth}
         \centering
         \includegraphics[width=\textwidth]{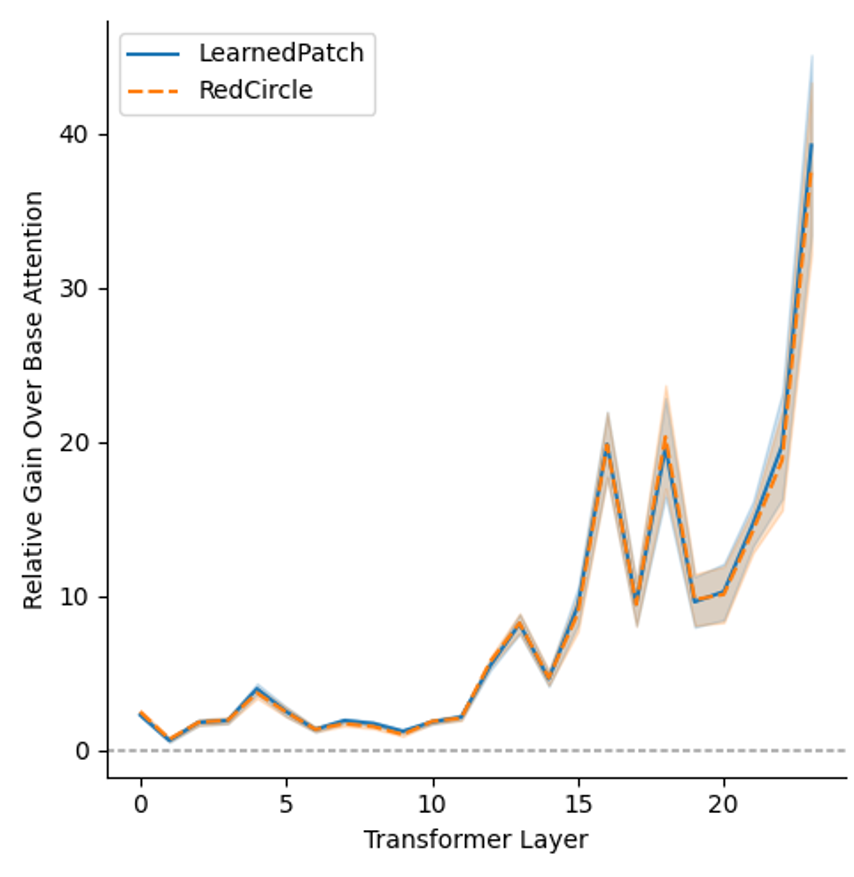}
         \caption{\clipl}
     \end{subfigure}
     \hfill
     \begin{subfigure}[b]{0.18\textwidth}
         \centering
         \includegraphics[width=\textwidth]{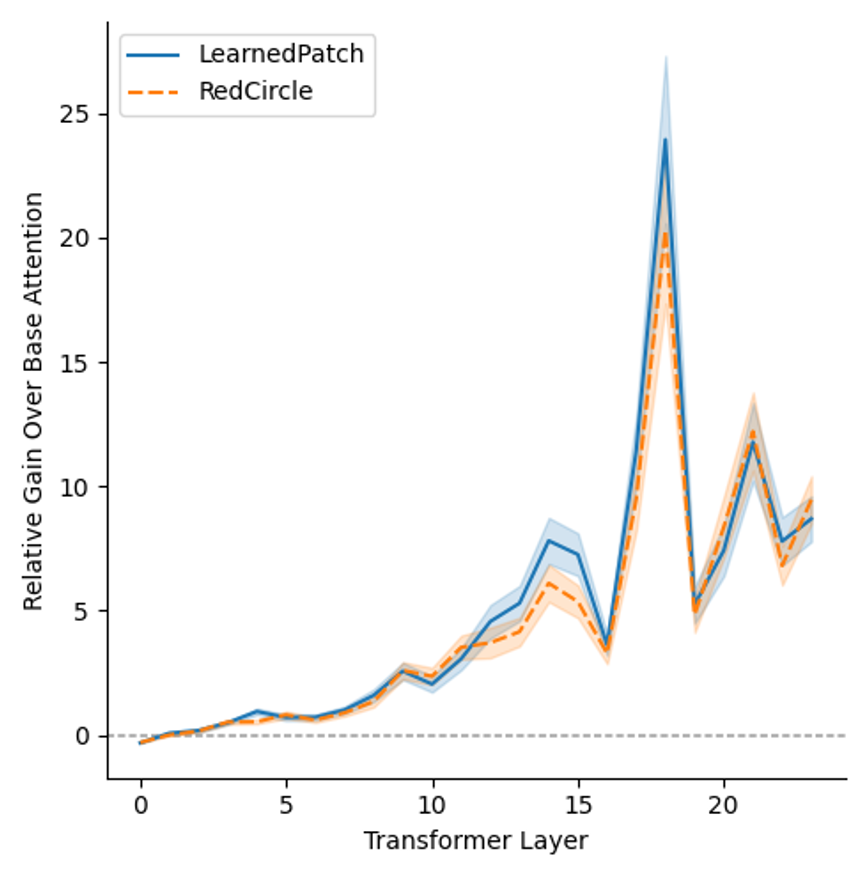}
         \caption{\siglip}
     \end{subfigure}
     \hfill
     \begin{subfigure}[b]{0.18\textwidth}
         \centering
         \includegraphics[width=\textwidth]{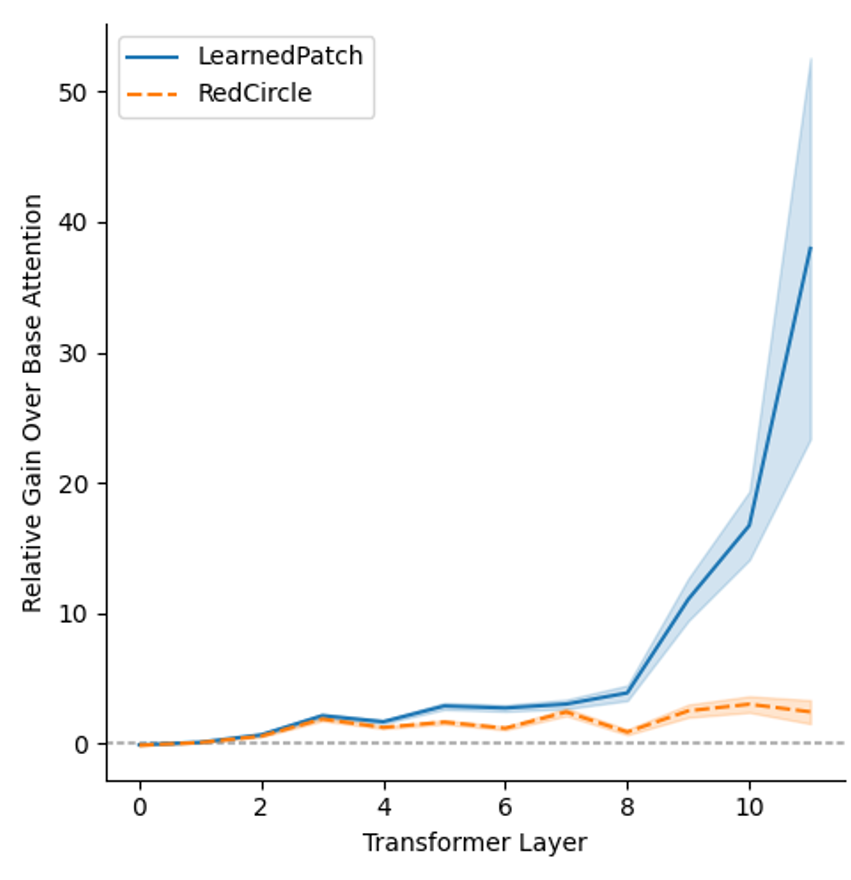}
         \caption{\deit}
     \end{subfigure}
     \hfill
     \begin{subfigure}[b]{0.18\textwidth}
         \centering
         \includegraphics[width=\textwidth]{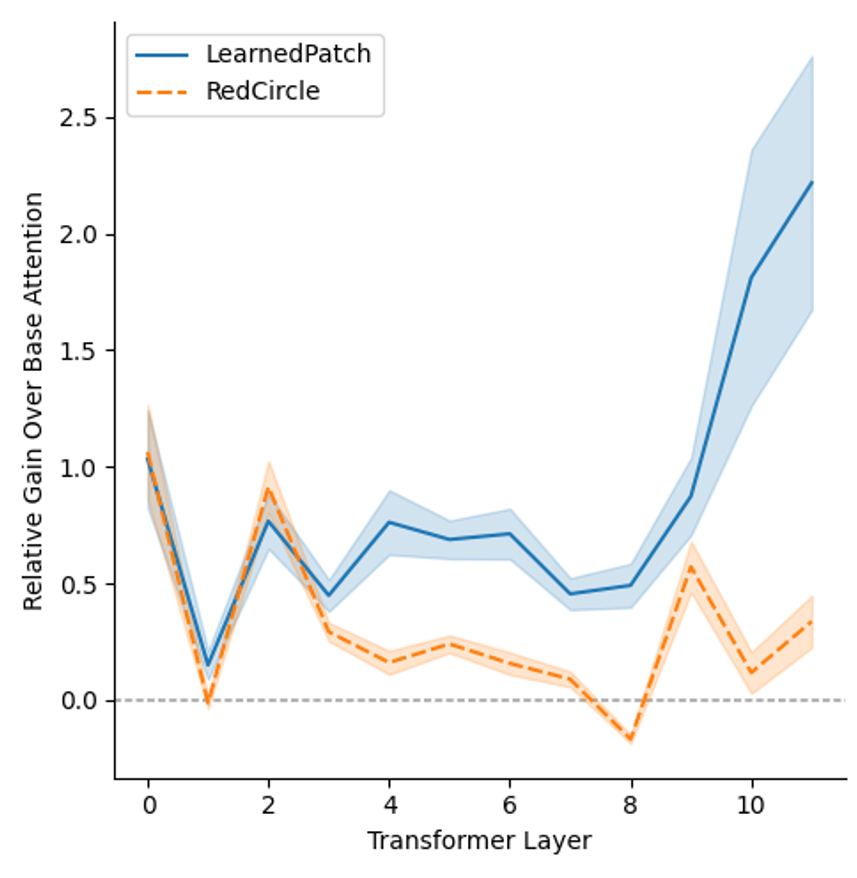}
         \caption{\dino}
     \end{subfigure}
  \caption{\textbf{Attention Gain with Prompt Usage throughout Layers.}
    We applied our learned prompt to random locations on 1000 samples from the MS COCO dataset (blue line). For comparison, we also used a simple red circle prompt on the same locations (orange line).
    The Relative Gains are calculated by dividing the difference between attention values before and after applying the prompt on the image by the original attention values of the overlaid tokens without any prompt. A notable observation is that while the simple red circle prompt is as effective for \clipl and \clipb, it is significantly less effective for \deit and \dino compared to the learned prompt in terms of redirecting attention to a specific location.
    }
    \label{fig:attGains}
\end{figure}

\textbf{The effect of the optimized prompt on attention across layers: }
We examine the attention averaged values of tokens in the area overlaid by the prompt before and after its application for over 1000 images from MS COCO while the prompt location is random. By analyzing attention values across layers, we can assess the prompt's impact. Since attention values are relative, we define Attention Gain as the ratio of the difference of averaged attention values of tokens influenced by the prompt to their original attention values. Figure \ref{fig:attGains} illustrates the attention gains across different models. The blue line represents the gains for each model's optimized prompt, while the orange line shows the gains for a simple red circle marker prompt.

\textbf{Red circle does not guide attention of \dino and \deit:}
From Figure \ref{fig:attGains} we can see that the unique prompts for \deit and \dino draw significantly larger attention to themselves compared to the red circle prompt, showing the power of our learning framework.
This could confirm the suggestion that the presence of red circles in the training data of CLIP models and \siglip has made them particularly sensitive to this predefined feature \cite{Shtedritski_2023_ICCV_RedCircle}, which is not the case for other vision encoders such as \deit and \dino.

\begin{figure*}[t]
\begin{center}
\includegraphics[width=\linewidth]{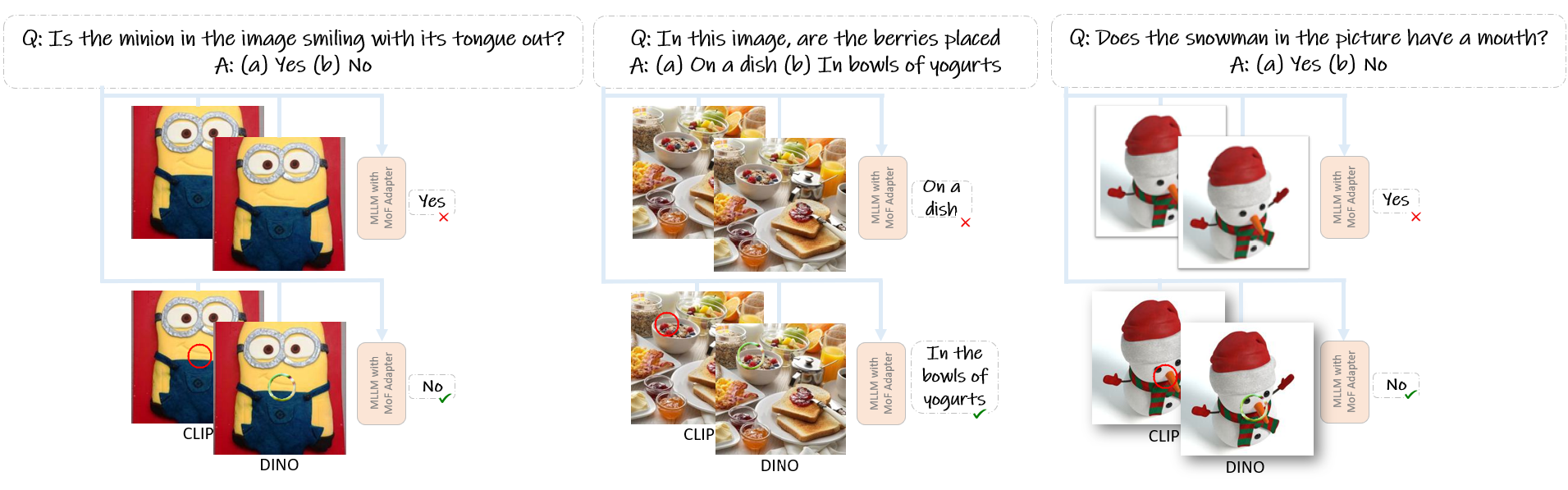}
\caption{\textbf{Prompts for the new generation of Vision Language Models:} New vision language models are moving beyond clip vision encoders towards using richer vision encoders such as \dino.
Therefore it becomes progressively necessary to identify visual prompts for the new generation of models.
This figure depicts an example to show the potential of using the optimized prompts in the new generation of vision-language models, in this case,
LLaVA+MoF \cite{tong2024eyesWide} on examples from the MMVP dataset.}
\label{fig:eyesWideShut}
\end{center}
\end{figure*}

\textbf{The significance of optimizing prompts for self-supervised models such as \dino:}
To further evaluate the effectiveness of our prompts, we use the MLLM (Multimodal Large Language Model) introduced in \cite{tong2024eyesWide}, which uses an \textit{Interleaved Mixture-of-Features} approach to spatially leverage interleaving CLIP and \dino visual tokens after an adapter.
In Figure \ref{fig:eyesWideShut}, we compare \cite{tong2024eyesWide} model's performance with and without our unique \clipl and \dino prompts applied to the image input to see if it improves the question answering responses of their model on proposed MMVP (Multimodal Visual Pattern) Benchmark.
The examples demonstrate that our prompts enable the model to generate more accurate answers, which means they create improved embeddings.
This indicates that our attention-guiding prompts have promising applications in a variety of vision tasks that can be explored in future works.

\section{Conclusion}
In this work, we proposed a self-supervised optimization-based visual prompting technique for guiding the attention of vision transformers, thereby avoiding the limitations of manually crafted prompts. Our method demonstrated the ability to guide the attention of various vision transformer models, such as the CLIP family, \siglip, \deit, and \dino, without requiring prior knowledge of dataset biases and without fine-tuning the models. The transferability of the prompt across different images was accomplished by leveraging a deep network prior. Our experiments confirmed that the learned prompts successfully redirect the attention of not only the CLIP family of models which are trained with language supervision, but also the purely self-supervised models such as \dino. As new vision language foundation models are leveraging the self-supervised vision encoders for their superior ability to extract visual features, the proposed optimization-based proves to be helpful for prompting upcoming models.

\bibliographystyle{abbrvnat}
\bibliography{mainbib}

\end{document}